\documentclass[conference]{IEEEtran}
\IEEEoverridecommandlockouts
\usepackage{cite}
\usepackage{amsmath,amssymb,amsfonts}
\usepackage{algorithmic}
\usepackage{graphicx}
\usepackage{textcomp}
\usepackage{xcolor}

\def\BibTeX{{\rm B\kern-.05em{\sc i\kern-.025em b}\kern-.08em
    T\kern-.1667em\lower.7ex\hbox{E}\kern-.125emX}}
\begin{document}

\title{Towards Explaining Autonomy with Verbalised Decision Tree States}

\author{\IEEEauthorblockN{Konstantinos Gavriilidis${^1}$, Andrea Munafo${^2}$, Helen Hastie${^1}$, Conlan Cesar${^3}$, Michael DeFilippo${^3}$, Michael R. Benjamin${^3}$}
\IEEEauthorblockA{\textit{${^1}$Edinburgh Centre for Robotics, Heriot-Watt University, Edinburgh, UK \{kg47, H.Hastie\}@hw.ac.uk}}
\IEEEauthorblockA{\textit{${^2}$SeeByte Ltd., Edinburgh, UK, andrea.munafo@seebyte.com}}
\IEEEauthorblockA{\textit{${^3}$Massachusetts Institute of Technology, Cambridge, Massachusetts, USA, \{conlanc, mikedefm, mikerb\}@csail.mit.edu}}
% \and
% \IEEEauthorblockN{Conlan Cesar, Michael DeFilippo, Michael R. Benjamin}
% \IEEEauthorblockA{\textit{Dept. of Mechanical Engineering} \\
% \textit{Massachusetts Institute of Technology}\\
% Cambridge, Massachusetts, USA \\
% \{conlanc, mikedefm mikeb\}@csail.mit.edu}
% \and
% \IEEEauthorblockN{Michael DeFilippo}
% \IEEEauthorblockA{\textit{Dept. of Mechanical Engineering} \\
% \textit{Massachusetts Institute of Technology}\\
% Cambridge, Massachusetts, USA \\
% mikedef@mit.edu \orcidlink{0000-0001-6214-0103}}
% \and
% \IEEEauthorblockN{Michael R. Benjamin}
% \IEEEauthorblockA{\textit{Dept. of Mechanical Engineering} \\
% \textit{Massachusetts Institute of Technology}\\
% Cambridge, Massachusetts, USA \\
% mikerb@csail.mit.edu \orcidlink{0000-0002-2520-6465}}
}

\maketitle

% \begin{IEEEkeywords}
% Explainable Autonomy, Natural Language Generation, Explainable AI, Robotics, Human-Robot Interaction
% \end{IEEEkeywords}

\section{Introduction}
The development of new AUV technology increased the range of tasks that AUVs can tackle and the length of their operations.
AUVs are today able to handle highly complex operations. However, these missions do not fit easily into the traditional method of defining a mission as a series of pre-planned waypoints because it is not possible to know, in advance, everything that might occur during the mission.
This results in a gap between operator's expectations and actual operational performance. This then can create a diminished level of trust between the operators and AUVs, which can in turn result in unnecessary mission interruptions.

\begin{figure}[t!]
\centering
\includegraphics[width=1.0\columnwidth]{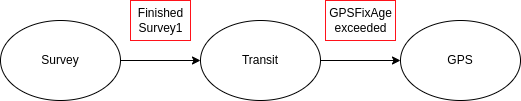}
\caption{Illustration of a behaviour chain together with the operational reason behind each activation.}
\label{fig: behaviour_chain}
\end{figure}

In behavioural autonomy\cite{b0, b1,b2}, multiple behaviours are available to allow the robot to adapt to any circumstance and complete a mission. 
In Figure \ref{fig: behaviour_chain}, for example, a simple sequence of behaviours is shown.
In this case, the robot initially uses a \textit{Survey} behaviour to explore an area, and once the objective \textit{Survey1} is complete, a \textit{Transit} behaviour is triggered to move to the next waypoint. During the transit, the vehicle could also trigger a \textit{GPS} behaviour to obtain an  GPS fix and update its position. 
This would temporarily interrupt its current task (transit), to move the vehicle to the surface until a new position update is received, at which point the transit behaviour can resume. If the completion of an objective or the use of updated thresholds are not communicated to the end user, an experienced operator of the system could think of a number of reasons for each transition. A user with limited or no experience might not even be able to make a guess. 
In underwater applications where communications are unreliable, limited in bandwidth and delayed~\cite{b3, b4} mission supervision becomes even more challenging: when unexpected behaviours occur there is typically only very limited or no information to explain it at the command and control level.

To bridge this gap between in-mission robotic behaviours and operators' expectations, this work aims to provide a framework to explain decisions and actions taken by an autonomous vehicle during the mission, in an easy to understand manner.
Additionally, the objective is to have an autonomy-agnostic system that can be added as an additional layer on top of any autonomy architecture. 
To make the approach applicable across different autonomous systems equipped with different autonomies, this work decouples the inner workings of the autonomy from the decision points and the resulting executed actions applying \textit{Knowledge Distillation}~\cite{b5}.

Knowledge distillation \cite{b5} makes it possible to interpret deterministic autonomous agents by building an equivalent representation, in our case, a distilled decision tree.
The decision tree acts as a mediator retrieving the vehicle state in real time, and based on that, generates the corresponding state-actions tree traversals that match the autonomy decision-making process with the highest probability. 

Finally, to present the explanations to the operators in a more natural way, the output of the distilled decision tree is combined with natural language explanations ~\cite{b6} and reported to the operators as sentences (see Figure~\ref{fig: pipeline_architecture}). For this reason, an additional step known as \textit{Concept2Text Generation}~\cite{b7} is added at the end of the explanation pipeline. 
More details on each step are given in the next sections.

\section{Decision Tree Distillation}
To distill relevant information and build an equivalent, distilled autonomy representation, the explanation pipeline uses sensor, mission and navigation data. In our implementation, this is done in two phases, firstly connecting a ROS node to listen to relevant topics (e.g., behaviour changes, nav data, etc.) and secondly learning simple decision rules inferred from the data features to then predict the value of the corresponding behaviour. In our case, tree splits represent state conditions for activating a behaviour, while leaf nodes represent the activated behaviours. For example:

\begin{figure*}[h!]
\centering
\includegraphics[width=0.7\textwidth,height=7cm]{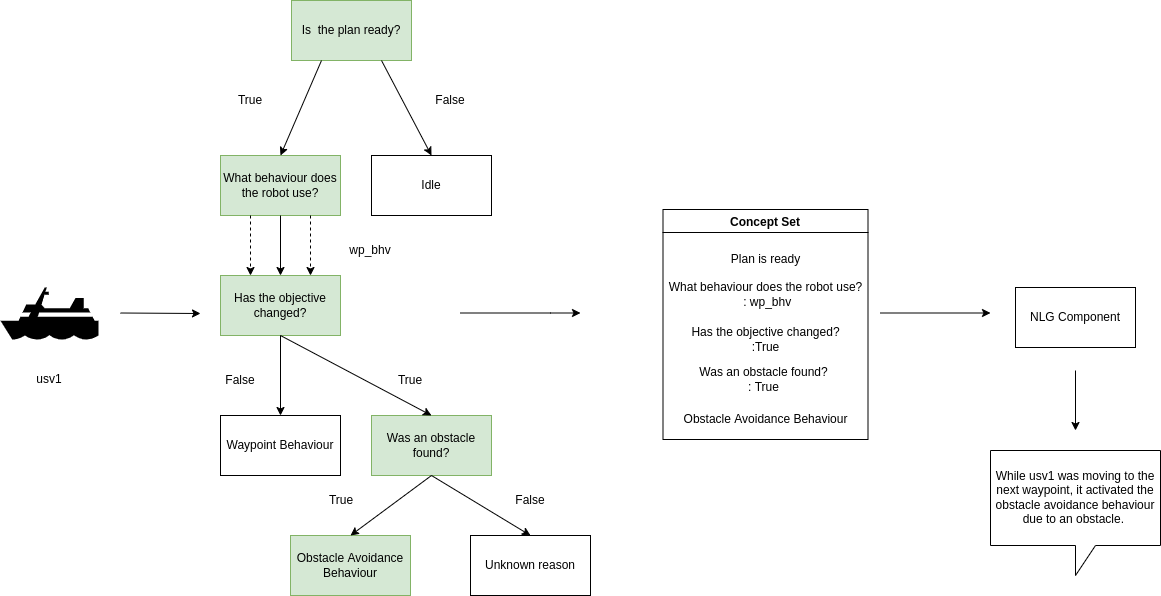}
\caption{Illustration of the proposed pipeline architecture with Knowledge Distillation and Concept2Text Generation using an NLG model. 
}
\label{fig: pipeline_architecture}
\end{figure*}

\[ behaviours = \{goto,transit,survey,wait,gps,avoid-obstacles\} \]
\[ states = \{battery, location, velocity, excl zones, objectives\} \]

Once a decision tree is available, it is used by the explanation pipeline to find the tree traversal that best matches the available data coming from the vehicle.

\section{Concept Set Extraction}
Each tree traversal is finally represented through a concept set, which includes: \textbf{(behaviour, causality, time)}.
Behaviour is the active vehicle behaviour, causality represents the state-action tree traversal, and time captures the passing of time, for example through timestamps.
Figure~\ref{fig: pipeline_architecture} represents an example of decision tree traversal, where a robot has a plan, which is under execution, and has activated its waypoint behaviour to get to the first waypoint of the mission.
The decision tree, in this case, is also able to represent planning changes due to the presence of obstacles.
In this case, when the decision tree detects a change in the current objective, is also able to capture a corresponding change in the activated behaviour.

The running behaviour changes to \textbf{Obstacle Avoidance} and the reason of its activation would be: \textbf{Was an obstacle found?}. The concept set output in this case is: \textbf{(obstacle avoidance, obstacle found, time)}. Subsequently, to generate explanations from the extracted concept sets, a number of NLG methods are being investigated \cite{b8}.

\begin{figure}[h!]
    \centerline{\includegraphics[width=6.0 cm]{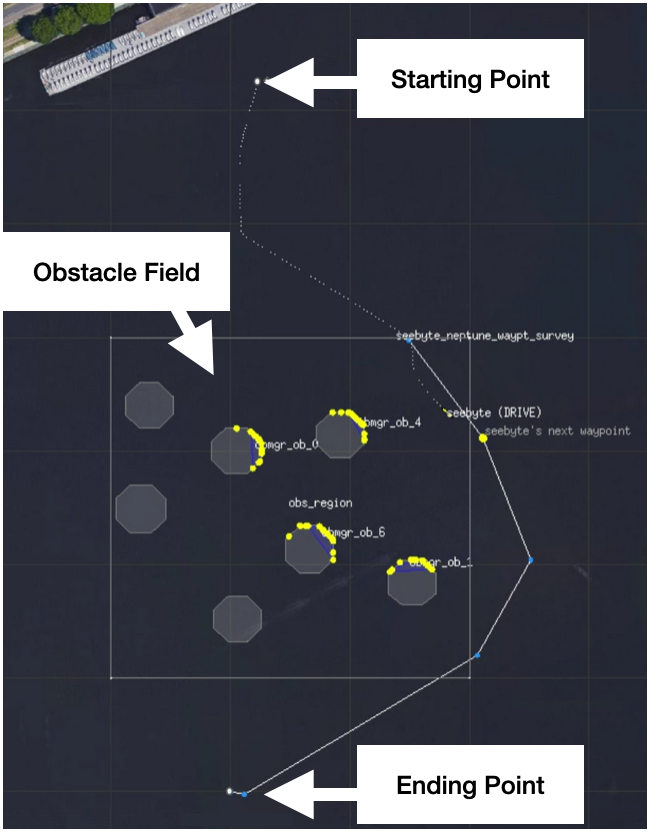}}
    \centering
    \caption{Simulation of Neptune displayed on MOOS-IvP pMarineViewer (see~\cite{b9}).}
    \label{fig:neptune-moos-sim1}
\end{figure}

\section{Simulations and Experiments}
The system has been tested in a multi-autonomy simulated environment where the Neptune autonomy was planning around an obstacle field~\cite{b9} (see Figure~\ref{fig:neptune-moos-sim1}). 
Initial results show that the decision trees are able to correctly capture the underlying autonomy and the relevant behaviour switching during the main decision events (i.e, obstacle detection), while the Concept2Text generation model provides sentences that represent the tree traversal.
To validate this initial simulated performance, the system will be tested during in-water activities planned for the summer 2022 in the Charles River in Boston, USA. 
The experiment plan includes one or more surface vehicle (USV) and one underwater vehicle (UUV). Following \cite{b9}, each vehicle is equipped with both SeeByte's Neptune Autonomy and MIT's MOOS-IvP coordinating their actions together to achieve the best mission outcome. The explanation pipeline, which will be located ashore, will use the data made available by the vehicles during the mission to provide explanations in real-time to operators.

\section*{Acknowledgment}
This work was funded and supported by the EPSRC Prosperity Partnership (EP/V05676X/1), the Office of Naval Research (ONR) under contract n. N00014-19-C-2056, the UKRI Node on Trust (EP/V026682/1), EPSRC CDT on Robotics and Autonomous Systems (EP/S023208/1), and SRPe.
%The authors want to express their gratitude to Michael R. Benjamin, Conlan Cesar and Michael DeFilippo for implementing the simulator and for the fruitful discussions.

\end{document}